\documentclass[pdflatex,sn-mathphys-num]{sn-jnl}

\usepackage{graphicx}%
\usepackage{multirow}%
\usepackage{amsmath,amssymb,amsfonts}%
\usepackage{amsthm}%
\usepackage{mathrsfs}%
\usepackage[title]{appendix}%
\usepackage[table]{xcolor}
\usepackage{textcomp}%
\usepackage{manyfoot}%
\usepackage{booktabs}%
\usepackage{algorithm}%
\usepackage{listings}
\usepackage{amsmath,amssymb,amsfonts}
\usepackage{algorithmic}
\usepackage{graphicx}
\usepackage{textcomp}
\usepackage{xcolor}
\usepackage{censor}
\usepackage{subcaption}
\usepackage{tabularx}

\begin{document}

\title{Adaptive Stream Processing on Edge Devices through Active Inference}

\author*[1]{\fnm{Boris} \sur{Sedlak}}
\email{b.sedlak@dsg.tuwien.ac.at}

\author[2]{\fnm{Victor} \sur{Casamayor Pujol}}
\email{v.casamayor@dsg.tuwien.ac.at}

\author[1]{\fnm{Andrea} \sur{Morichetta}}
\email{a.morichetta@dsg.tuwien.ac.at}

\author[3]{\fnm{Praveen Kumar} \sur{Donta}}
\email{praveen@dsv.su.se}

\author[1,2]{\fnm{Schahram} \sur{Dustdar}}
\email{dustdar@dsg.tuwien.ac.at}

\affil[1]{\orgdiv{Distributed Systems Group}, \orgname{TU Wien}, \orgaddress{\city{Vienna}, \postcode{1040}, \country{Austria}}}

\affil[2]{\orgdiv{Distributed Intelligence and Systems-Engineering Lab (DISL)}, \orgname{Universitat Pompeu Fabra}, \orgaddress{\city{Barcelona}, \postcode{08018}, \country{Spain}}}

\affil[3]{\orgdiv{Department of Computer and Systems Sciences}, \orgname{Stockholm University}, \orgaddress{\city{Stockholm}, \postcode{16425}, \country{Sweden}}}


\abstract{
% \rw{Every year, the amount of data created by Internet of Things (IoT) devices increases; therefore, data processing is carried out by edge devices in close proximity. To ensure Quality of Service (QoS) throughout these operations, systems are supervised and adapted with the help of Machine Learning (ML). However, as long as ML models are not retrained, they fail to capture gradual shifts in the variable distribution, leading to an inaccurate view of the system state and poor inference. In this paper, we present a novel ML paradigm that is constructed upon Active Inference (AIF) -- a concept from neuroscience that describes how the brain constantly predicts and evaluates sensory information to decrease long-term surprise. We implemented a use case, in which an AIF-based agent continuously optimized the operation on a smart manufacturing engine according to QoS requirements. The agent used causal knowledge to gradually develop an understanding of how its actions are related to requirements fulfillment, and which configurations to favor. As a result, our agent required five cycles to converge to the optimal solution.
% }
The current scenario of IoT is witnessing a constant increase on the volume of data, which is generated in constant stream, calling for novel architectural and logical solutions for processing it. Moving the data handling towards the edge of the computing spectrum guarantees better distribution of load and, in principle, lower latency and better privacy. However, managing such a structure is complex, especially when requirements, also referred to Service Level Objectives (SLOs), specified by applications' owners and infrastructure managers need to be ensured. Despite the rich number of proposals of Machine Learning (ML) based management solutions, researchers and practitioners yet struggle to guarantee long-term prediction and control, and accurate troubleshooting. Therefore, we present a novel ML paradigm based on Active Inference (AIF) -- a concept from neuroscience that describes how the brain constantly predicts and evaluates sensory information to decrease long-term surprise. We implement it and evaluate it in a heterogeneous real stream processing use case, where an AIF-based agent continuously optimizes the fulfillment of three SLOs for three autonomous driving services running on multiple devices. The agent used causal knowledge to gradually develop an understanding of how its actions are related to requirements fulfillment, and which configurations to favor. Through this approach, our agent requires up to thirty iterations to converge to the optimal solution, showing the capability of offering accurate results in a short amount of time. Furthermore, thanks to AIF and its causal structures, our method guarantees full transparency on the decision making, making the interpretation of the results and the troubleshooting effortless.}


\maketitle

\section{Introduction}

% \ins{must explain that this is an extension of \cite{sedlak_active_2024}}

% \ins{must emphasize the resource constraints further and how our methodology can provide elasticity on edge devices; also how ML is usually black box and we're not}

Recent years have reported a constant transition of logic and computation from the central cloud towards the edge of the network \cite{deng_edge_2020}, bringing the execution closer to the Internet of Things (IoT) devices that actually generate data. The reasons behind migrating the computation to the edge span from saving bandwidth, to improving privacy,
%~\cite{bekbulatova2023fl}, 
as well as speeding up data processing (i.e., decreasing latency). %\cite{pujol2023edge}. 
Furthermore, this transition, facilitated by improvements in hardware and connectivity, currently allows the training and deployment
of Machine Learning (ML), up to Deep Learning (DL) models. This step is pivotal in such scenarios, as ML models play an essential role in interpreting and predicting the behavior of applications and distributed systems, in order to guarantee Quality of Service (QoS). Indeed, in the last years ML has been vastly utilized~\cite{hua2023edge} for the management of complex, distributed infrastructures and applications, with tasks that go from estimating the impact of redeployment \cite{chen_causeinfer_2019} to forecasting potential system failures \cite{morichetta_demystifying_2023}.%, which must be circumvented to ensure the Quality of Service (QoS).

However, contrarily to the cloud infrastructure, deploying services on top of resource-restricted edge devices while guaranteeing their QoS is not trivial. Indeed, edge devices, which span from commodity servers to dedicated machines, have a constrained pool of resources. Typically, a set of common strategies involve letting edge devices scale their services through local reconfiguration, assisted by ML \cite{furst_elastic_2018}. Despite the validity of these strategies, most of them do not consider when the initial conditions changed. This is especially limiting for ML models, which are not retrained although new observations would be available \cite{morichetta_demystifying_2023,chen_causeinfer_2019}, inevitably leads to a drift in observing and predicting the system state. Exemplifying it, imagine an elastic computing system, as envisioned in \cite{nastic_sloc_2020}, which observes the system through a set of metrics, evaluates whether QoS requirements -- also called Service Level Objectives (SLOs) -- are fulfilled, and dynamically reconfigures the system to ensure SLOs are met. If the variable distribution changes and the ML model is not adjusted, this makes it impossible to interpret system metrics correctly, and any consequential reconfiguration will fail to fulfill its purpose.
Therefore, a solution to guarantee the precision of ML models over time involves implementing continuous feedback mechanisms; this could, for example, be achieved by optimizing a value function, as in reinforcement learning (RL)~\cite{martinez_probabilistic_2021,friston_reinforcement_2009}. However, approaches like RL are slow to converge and typically computationally intensive. Furthermore, leveraging ML algorithms means, nowadays, often relying on DL models. Despite their great performance in various, complex applications, most of them suffer from a lack of interpretability, which is instead essential in core tasks like distributed infrastructure and application management. For these reasons, we believe that this scenario requires a more holistic approach, which starts with making the SLOs first-class citizens during ML training. Further, any component that uses ML for inference should actively resolve or report ambiguities. We envision that such a level of self-determination could be provided by Active Inference (AIF), a concept from neuroscience that describes how the brain constantly predicts and evaluates sensory information to decrease long-term surprise. In cases where ML training and inference are carried out in close proximity to the data source, i.e., on edge devices, AIF can ensure model accuracy whenever the accuracy drops. Equipped with AIF, edge devices could continuously infer system configurations that ensure QoS. Furthermore, AIF allows to develop causal understanding of a process; this raises the trust for inferred results \cite{chen_causeinfer_2019,sedlak_designing_2023}.

In this paper, we extend our previous contribution~\cite{sedlak_active_2024}, where we presented a design study of an AIF agent optimizing the throughput in a smart factory. To show the general applicability of our approach, we now apply AIF extensively for distributed processing systems and provide a novel evaluation with 3 Edge-based stream services showing how the agents are able to adapt service configuration to match QoS requirements.
Agents operate autonomously and decentralized while ensuring the SLO compliance on their local edge devices. At its core, the agent follows an action-perception cycle where it first estimates which parameter assignments would violate given SLOs, then compares this expectation with new observations, and finally, adjusts its beliefs (i.e., the ML model) accordingly.
While exploring the value space, it favors solutions that are likely to improve the model precision; this, in turn, provides the agent with a clear understanding of the causal relations between model variables. Hence, the contributions of this article are:

\begin{itemize}
    \item An adaptive stream processing mechanism based on Active Inference that continuously optimizes the QoS of streaming pipelines through local reconfiguration; thus, processing services scale autonomously according to environmental impacts
    % \item A novel ML paradigm based on AIF that continuously evaluates the quality of inferred configurations. Thus, edge devices maintain QoS requirements fulfilled.
    \item A transparent decision-making process for AIF agents that adjusts processing configurations according to expected SLO fulfillment and model improvement. The underlying causal structures make results empirically verifiable and increase trust.
    % \item The composite representation of agents' behavior according to causal relations and empirical information. This increases trust and reproducibility of inferred results.
    \item The evaluation of the presented AIF agent under three different stream processing services and heterogeneous edge device types. This underlines how the presented methodology is apt to support a wider range of stream processing scenarios.
    % \item A complete design study for a smart manufacturing use case that paves the way for other researchers to implement AIF in related automotive use cases.
    
\end{itemize}

The remainder of the paper is structured as follows: Section~\ref{sec:background} provides background information on AIF principles in edge computing; Section~\ref{sec:related-work} presents related work; in Section~\ref{sec:agent-design} we outline the design process of an AIF agent, which we implement and evaluate in Section~\ref{sec:analysis}. Finally, Section~\ref{sec:conclusion} concludes the paper.

% \newpage

\section{Background}
\label{sec:background}

To dynamically adjust stream processing, the core mechanism applied in this paper is AIF; given that AIF is a concept that originates from neuroscience, we use this section to summarize core concepts of AIF according to Friston et al.~\cite{friston_life_2013,kirchhoff_markov_2018,parr_active_2022,friston_designing_2024}. This includes (1) a high-level picture of how AIF works, (2) an illustrative example of how agents use AIF, and (3) a formal representation of how AIF agents adjust their beliefs and choose actions that fulfill their preferences. Following that, we delineate our view of the intersection between AIF and distributed computing systems, highlighting how AIF concepts are relevant for stream processing on edge devices.

\subsection{Free Energy Principle and Active Inference}
\label{subsec:FEP-and-AIF}

The Free Energy Principle (FEP) and its corollary Active Inference (AIF) stem from neuroscience, aiming at answering the overarching question: \emph{how does the brain work?} 

\vspace{5pt}
\textbf{The high-level picture.} 
Seeking to answer such a fundamental and complex question, Friston offers a theory~\cite{friston2010free} of an overarching structure for the brain and cognition. In their understanding, cognitive agents have the capacity to build an \textit{internal model} of their observed environment and the underlying \textit{generative processes}. This internal model -- also called a \textit{generative model} -- is used by agents to \textit{predict and adjust} their environment according to their \textit{preferences}. In the FEP theory, and especially in AIF, the system revolves around two core elements: the \textit{generative process} and the \textit{generative model}. The first is the underlying causal structure of the environment's behavior that produces agent's observations; the \textit{generative model} approximates, from the agent's perspective, the behavior of the environment.

Using their generative model, agents predict and modify their environment; for this, they assume that the generative model aligns closely (i.e., in terms of KL divergence) with the generative process.
However, if the generative process and the model differ, this discrepancy will ``surprise'' the agent, causing it to adjust its model closer towards the process~\cite{bruineberg2018free}. This surprise (after Bayesian surprise~\cite{itti2009bayesian}) plays a fundamental part in Friston's theoretical framework because \textit{Free Energy (FE)} formally presents an upper bound on surprise.
To minimize FE, and hence surprise, AIF agents constantly engage in action-perception cycles, where they (1) predict sensory inputs, observe the environment, and update their beliefs depending on the outcome. % -- widely known as predictive coding. 
Afterward, they (2) actively adjust the world to their preferences. 
% (ii)
Internally, agents organize their generative models in hierarchical structures; each level interprets lower-level causes and, based on that, provides predictions to higher levels. This process of using existing beliefs (widely known as priors) to calculate the probability of related events is commonly known as Bayesian inference~\cite{pearl_causal_2009} and it allows agents to improve their understanding of the environment. AIF models are structured by self-contained Bayesian structures, known as Markov Blankets (MBs), that separate the internal states of a system from its outside environment, formalizing the action-perception cycle.

% (iii)

\vspace{5pt}
\textbf{An illustrative example.}
To exemplify this approach, imagine an individual who believes that it is raining; as a consequence, he will take an umbrella to avoid getting wet, as this would otherwise be uncomfortable.
As defined before, individual agents interpret observable processes through generative models. In this case, an individual might reason that it is raining as he observes water drops falling from the sky. 
% By observing the environment, the agent can learn to understand real-world processes.
Using its generative model, the individual infers with high probability that it is raining; the respective action is to take an umbrella to fulfill its preference of staying dry.
However, imagine that the observed water drops were actually caused by a neighbor watering their plants. Thus, the generative model and the process diverge, and the agent is ``surprised" once he leaves the house.
More generally, individuals can either take pragmatic action, e.g., picking an umbrella to fulfill their preferences, or otherwise, improve decision-making by exploring the environment through epistemic actions. For example, looking at the blue sky and the neighbor's balcony reveals that drops resulted from watering the plants, avoiding the later consequence of carrying an umbrella on a sunny day. The agent thus updates its prior beliefs (i.e., rain $\rightarrow$ water) according to new information (i.e., rain $\rightarrow$ water $\leftarrow$ flowers) to form its posterior beliefs.

% \ins{put an example graph for the simple example here with the rain}

% \ins{change the illustrative example to something more related}

\vspace{5pt}
\textbf{Formalization.}
Picking up from the explanations of AIF and FE, it remains to provide a formal representation of these concepts: given that an agent equipped with generative model $m$ makes an observation $o$, the surprise $\Im (o|m)$, as shown in Eq.~\eqref{eq:surprise}, is the negative log-likelihood of the observation~\cite{parr_active_2022}.
The FE of the model -- expressed as the Kullback-Leibler divergence ($K\!L$) between approximate posterior probability ($Q$) of hidden states ($x$) and their exact posterior probability ($P$) -- is an upper-bound on surprise. This is formalized in Eq.~\eqref{eq:free-energy}; however, the exact posterior probability ($P$) is practically intractable, hence, $Q$ is used as an approximation. 
% With the exact posterior approximated by $Q$,
The FEP now uses variational inference~\cite{blei_variational_2017} as the method to find the best approximation that minimizes the difference between both distributions.

\begin{equation}
    \Im (o|m)= -\ln\!\!{\overbrace{P(o|m)}^\text{Model Evidence}}
    \label{eq:surprise}
\end{equation}
\vspace{-3pt}
\begin{equation}
    F[Q,o] = \underbrace{K\!L[Q(x)||P(x|o,m)] + \Im (o|m)}_\textrm{(Variational) Free Energy} \geq \Im (o|m) 
    \label{eq:free-energy}
\end{equation}
\vspace{5pt}

AIF agents use this mathematical framework to select among actions~\cite{smith_step-by-step_2022, parr_active_2022}; for this, they condition all terms on the policy ($\pi$). However, instead of looking into the approximate model performance, AIF agents select the best policy according to possible future states, actions, and observations. More formally, agents minimize their Expected Free Energy (EFE) by selecting the optimal policy. The EFE can be expressed as in Eq~\eqref{eq:EFE}: 
\begin{equation}
\label{eq:EFE}
    EFE = -\overbrace{E_{Q(x|\pi)}[ \ln{P(x|R)} ]}^{\text{pragmatic value}} \, -\overbrace{E_{Q(x,z|\pi)} [ \ln{Q(z|x,\pi)} - \ln{Q(z|\pi)}]}^{\text{information gain}}
\end{equation}
where the first term, i.e., the pragmatic value (\textit{pv}), expresses how likely an action will produce observations that fulfill an agent's preferences (R). The \textit{pv} pushes an agent to take actions that will satisfy its goal; for a computing system, this means fulfilling its SLOs. Noteworthy, preferred observations can be understood as a reward function in reinforcement learning~\cite{friston_reinforcement_2009,tschantz_reinforcement_2020, sajid_active_2021}.
% Hence, the desired observations for the agents, i.e., fulfilling SLOs, have a higher weight in the preferred observations (C). 
The second term, i.e., the information gain (\textit{ig}), pushes an agent towards actions that allow it to learn about its environment; in the literature, \textit{ig} is also called ``epistemic value".
This expression shows that minimizing EFE intrinsically deals with the classic exploration-exploitation trade-off.

\subsection{AIF Principles in Distributed Systems}
\label{subsec:AIF-principles}

%\ins{maybe we can slightly rewrite this to point it more toward stream processing?}

Large-scale distributed computing systems, especially when handling streams of data and events coming from a wide range of devices and applications, consist of many interconnected components; therefore, modeling and managing~\cite{firmani2024intend, morichetta2024cohabitation} such structures requires ingenuity. In this regard, we identify three central challenges of distributed systems and stream processing, which can benefit from AIF concepts. 
First, (1) how to model large-scale systems and underlying structures to understand their overall functionality, e.g., to find the root cause of a bottleneck in the data processing pipeline.
Second, (2) how to mitigate SLO violations within distributed systems under unexpected runtime dynamics; this step is essential when we manage real-time, latency-sensitive scenario like in stream processing. All in all, the main goal boils down to taking the best actions at a given point in time.
% , when defining a static central strategy is not possible.
Finally, (3) how to maintain long-term fulfillment of the system's requirements through effective management; this step implies taking care of variations in the infrastructure and in the behavior and quality of data.
While there exist several ML-based methods for distributed computing management~\cite{ilager2020artificial}, most of these approaches leave several of these challenges still open.
In the following, we align these challenges with the main characteristics of AIF, elucidating how AIF can fundamentally improve the management of distributed computing systems.

\subsubsection{System Modeling through Causal Structures}
Understanding the behavior of distributed computing systems and reasoning on the effects of individual actions is challenging due to the entanglement and complexity among individual components. For example, finding the root cause of abnormal behaviors or backtracing which configuration decision in which part of the system propagated a certain system state is not trivial and is often computationally intractable. In this scenario, using causal models in distributed systems~\cite{pujol_causality_2024} is a captivating solution that can help identify the source of specific behaviors.

In this direction, a tool commonly used is Bayesian Networks~\cite{pearl_probabilistic_1988} (BN). It can be trained to identify connections and dependencies between components through conditional probability functions. While BNs offer a good understanding of correlation within a system, as Pearl points out~\cite{pearl_causal_2009,pearl_book_2018}, only observational data (i.e., data that we collect from a system without knowing what is its current state) is not enough to gain causal knowledge. Here, AIF plays an important role: when AIF agents start intervening in their environment, the agents gain more valuable insights into the system behavior. Repeating this action-perception cycle can enrich BN, gradually adding information that turns them into actual causal graphs.

In contrast to solely relying on Deep Learning methods, introducing causal structures to management solutions has the fundamental advantage of guaranteeing an interpretation of system behavior and the effects of management actions or recommendations, thus improving trustworthiness~\cite{ganguly_review_2023}. Specifically, in distributed computing systems, a causal structure can explain how metrics (e.g., latency or CPU load) are related to the system state~\cite{sedlak_designing_2023}, backtrack which service or device caused a system failure~\cite{chen_causeinfer_2019}, or predict the impact of redeployment~\cite{tariq_answering_2008}.

\subsubsection{Mitigating SLO Violations through Free Energy Minimization}

Modeling SLO fulfillment through BN or causal graphs, as explained above, is challenging. Replicating the probability distribution of events in large-scale systems is, at the very least, complex.
Hence, performing statistical inference is often intractable due to the distribution's complexity. Therefore, as shown in Eq~\eqref{eq:free-energy}, AIF leverages variational inference (VI)~\cite{blei_variational_2017}: VI uses the evidence lower bound (ELBO) to make the problem an optimization one. Without entering into details, increasing ELBO minimizes KL divergence and, therefore, allows to approximate the probability distribution. This problem can obviously be solved together with gradient methods and mechanisms like Deep Neural Networks. Indeed, ELBO and its variants are very popular in ML and DL (a notable example being Variational Auto Encoders~\cite{hoffman2016elbo}).

Minimizing KL-Divergence is crucial in AIF; this means, that agents intervening in the environment can lead to a better understanding of the environment and, as a consequence, makes it more likely to achieve agents' preferences. Naturally, illustrating this process inevitably draws a parallel with Reinforcement Learning (RL); however, AIF presents fundamental differences~\cite{sajid_active_2021}: whereas in RL, agents maximize a reward function, in AIF the agent intervenes in the environment to minimize EFE. This difference, albeit subtle, offers essential advantages. Minimizing EFE means not only performing the action that leads to observing the desired output (pragmatic value) but also improving the precision of the generative model (information gain).

In distributed computing systems, epistemic actions often suffice to reduce uncertainties about expected outcomes: distributed systems can resolve contextual information by identifying a low-utilized agent for tasks offloading \cite{huang_vehicle_2020,guo_toward_2020}, or evaluating resource availability before scaling a system \cite{sedlak_controlling_2023,furst_elastic_2018}. In other cases, taking action aids to prevent or mitigate SLO violations, e.g., in case of resource allocation or task offloading~\cite{wang2019smart, tang2020deep, huang2023joint, ju2023joint}. This embraces the common tradeoff between seeking either pragmatic value (exploitation) or epistemic value (exploration). Multi-agent systems~\cite{levchuk_active_2019} control this through hyperparameters, which foster early exploration of a value space but decay over time as agents report little improvement.

\subsubsection{Long-term System Balance through Homeostasis}

The ultimate goal for an AIF agent is to constantly minimize the surprise and thus persist over time; this requires guaranteeing that certain internal variables remain within defined ranges. In cybernetic theories of living organisms, this process is called \textit{homeostasis}. For example, the human body requires a core temperature of approximately 37° to ensure internal chemical processes. Hence, modeling the environment accurately allows agents to make the right decisions that fulfill its requirements, e.g., picking an umbrella when it's raining so that the body does not cool down.
% The AIF concept of ``surprise'' plays a significant role here, calling for dynamic adaptation mechanisms.
% In this case, the focus is on preemptively control the intervention in an allostatic fashion before interoceptive prediction errors are triggered. In the case of the human body, the individual can take a cold shower before overheating. the preferred strategy is to engage with the environment to correct this instead of changing the perception.

The human body, as a complex system, has distinct mechanisms to ensure internal requirements, e.g., in case of cooling down, it can raise the core temperature through shivering. The fact that shivering happens unconsciously underlines how systems, including distributed processing systems, benefit from autonomous requirements assurance. While processing systems can also pose requirements in terms of (CPU) temperature, these and other requirements are specified as Service Level Objectives (SLOs) that must be assured during processing. Common instances of SLOs are QoS requirements, such as response time or availability~\cite{nastic_sloc_2020}; in case SLOs are violated, processing systems can resolve this through elasticity strategies~\cite{lu_qos-aware_2023,zhang_octopus_2023}, e.g., cloud computing scales computational resources to limit response time. 
% own temperature-controlling mechanisms, distributed computing systems, enforce QoS requirements through SLOs \cite{sedlak_equilibrium_2024,sedlak_markov_2024}. More specifically, it is common to employ elasticity strategies to ensure QoS, e.g., by scaling computational resources to limit response time. 
Complex systems can have multiple elasticity strategies at their disposal~\cite{sedlak_controlling_2023}, e.g., computing systems generally scale three elasticity dimensions: resources, quality, and cost. To optimize SLO fulfillment through local reconfiguration, existing works~\cite{sedlak_designing_2023,sedlak_equilibrium_2024} used the notation of Markov blankets, which encompass relations between system variables and elasticity strategies. Thus, the agent can infer how to best ensure its requirements.
Simply put, an agent's preferred observation is having SLOs fulfilled; in case SLOs are violated, the agent will take action to correct this by choosing an adequate strategy according to the Markov blanket model.

\section{Related Work}
\label{sec:related-work}

%\ins{we still must highlight the research gap a bit better}
In this section, we delineate the state of the art in adaptive stream processing and its relation to AIF. First, we depict the current scenario in AIF, where, despite the praiseworthy efforts to theorize the framework, only a few contributions focus on implementing it, and still not in interdisciplinary scenarios. Later, we offer a snapshot of the prevailing approaches for adaptive stream processing and how approaches based on AIF can improve the QoS during stream processing.

\paragraph{AIF Applications}
While, to the best of our knowledge, there exists no off-the-shelf implementation of AIF in distributed systems; a handful of research works have combined AIF with computer science: 

The authors in \cite{vilas_active_2022} discuss AIF as a general computational framework, highlighting how existing research used AIF for (simulating) sensory processing. Touching on the design of AIF agents, Heins et al.~\cite{heins_pymdp_2022} provide a Python simulation that exemplifies how to structure action-perception cycles. Heins et al. further remark that existing AIF research largely focuses on formally constructing models in isolated environments~\cite{smith_step-by-step_2022} such as Matlab SPM rather than putting them into action, e.g., to improve the precision of ML models. Thus, a more hands-on application of AIF is to extend reinforcement learning with AIF principles
\cite{martinez_probabilistic_2021,friston_reinforcement_2009}. However, most research to date either uses only a few AIF principles or is not applied enough to easily transfer presented concepts to distributed systems.

The work in \cite{levchuk_active_2019} is, therefore, an exception because it embeds AIF into the IoT and describes how AIF can improve the behavior of adaptive agents. Thus, individual agents may dynamically regroup into hierarchical teams, federate knowledge, and collectively strive after a common goal (i.e., a search task). By emphasizing the information exchange between agents, they were able to speed up the convergence of the distributed task. However, while they focused on FE minimization, they did not treat the other two principles we identified for AIF in distributed systems: causal inference and homeostasis. In this paper, we will present an agent that uses all three AIF principles to infer actions, maintain agents' internal equilibrium, and persist over time. 
% Nevertheless, we will use the representation from \cite{levchuk_active_2019} for FE minimization.
% \\

\paragraph{Runtime adaptation of stream services}
Runtime adaptation of stream services is an extensive topic in the scientific literature. We narrow down the related work by leveraging the taxonomy of adaptation mechanisms presented by Cardellini et al.~\cite{cardellini_run-time_2021}. In that regard, we will focus on works that perform processing adaptations.
The initial driver for that boosted research related with the adaptation of stream services stems from commercial large-scale video streaming services that emerged more than a decade ago. For instance, Huang et al.~\cite{huang_buffer-based_2014} showed how to achieved 10-20\% less rebuffering rate while achieving a higher video rate.
Conversely to this research, our work focuses on computing paradigms outside the Cloud, aiming at adapting services that are at the Edge, next to IoT devices.

Leaning on more recent research, Khani et al.~\cite{khani_real-time_2021} presents a similar use case to ours, i.e., real-time video inference on Edge devices, but adapts the machine learning models deployed at the Edge devices by distilling knowledge from a centralized model. Further, they use a server, which contains the centralized model, to also adapt the frame sampling rate for each device. In contrast, our work aims toward a completely decentralized architecture, providing the Edge service the full autonomy to adapt.

Ma et al.~\cite{ma_edge_2022} develop a system to adapt video bitrate for live video streaming at the Edge based on Quality-of-Experience (QoE) requirements. They use deep reinforcement learning (DRL) to adapt the bitrate and distribute the video traffic to the clients according to their QoE needs. Similarly, Cao et al.~\cite{cao_neural_2023} build a DRL-based system to adjust video streams for Edge computing according to the expected QoE. Interestingly, they develop a programming model to facilitate the potential adaptations of the streams. 
% In distinction from these works, our approach based on SLOs opens the scope of QoE to also Quality of Service requirements.
Further, as discussed by Cao et al.~\cite{cao_neural_2023}, DRL models require periodical re-training as the data shifts while AIF includes this re-training within its normal formulation thanks to the possibility of performing epistemic adaptations. 

Similarly, as described by Furst et al.~\cite{furst_elastic_2018}, our work consists of achieving elastic services for stream processing at the Edge. Further we use SLOs to autonomously adapt each service. In contrast to Furst et al., our approach uses AIF to adapt stream processing services, which is a more flexible perspective for dynamic and long-lasting operations because AIF improves its behavior over time. 

\paragraph{Takeaways}
Despite existing contributions in the field of stream adaptation, we argue that our approach can offer better control in a more generalizable way. First, conversely, to previous stream services, we focus on defining ways to map SLOs for the autonomous control of a distributed computing system. Secondly, we manage that using AIF, which provides the flexibility for managing services that dynamically change over time. Finally, with our work, we contribute to the AIF field by both providing an implementation of AIF strategies and showing its potential in a real use case.

\section{AIF Adaptive Streaming Agent}
\label{sec:agent-design}

% \ins{After finishing S4, we will have to align with S2 to (1) avoid redundancy in S4 and (2) ensure that S2 contains all required concepts}

% \ins{present the high-level goals of the methodology: how it observes streaming processes, how it constructed generative models, and how these are applied by the action-perception agent}

In this Section, we delineate the three main steps for assembling an adaptive AIF agent for stream processing: (1) setting up tools and methods for observing the behavior of a continuous stream; (2) based on these outputs, training a generative model to predict and interpret stream processing behaviors; (3) implement the agent's algorithm for executing the AIF cycle, i.e., the application of the generative model for both action and perception. These three steps form a coherent, generalizable pipeline, applicable to different streaming use cases.

\subsection{Continuous Stream Processing}

% \ins{Gives abstract definitions of input data, output data, batches of observations}

\begin{figure}[t]
\centerline{\includegraphics[width=0.95\textwidth]{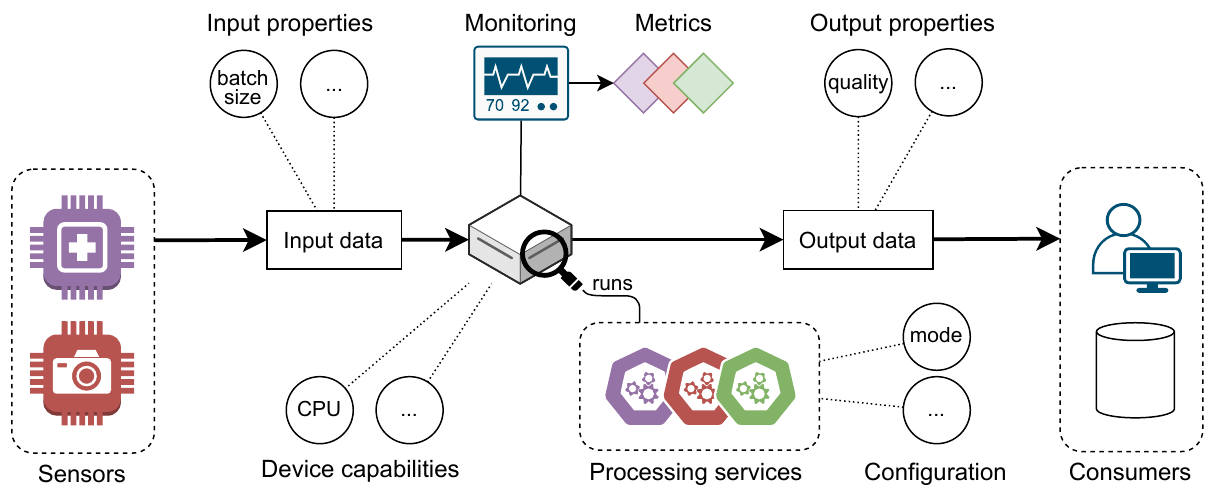}}\vspace{0.7em}
\caption{Abstract representation of a continuous stream processing scenario; sensors provide input data that is processed by services located at an edge device; processing is observable through metrics and respective results are provided to stream consumers}
\label{fig:stream-processing}
\end{figure}

Modeling a streaming service requires a well-defined architecture; first, there is the need of a transparent view of its internal processes, including the characteristics of the in- and out-flow data, e.g., batch size or data type in input, and quality of the processed output. Additionally, it is essential to have visibility on quantifiable metrics about the fulfillment of requirements.   %Streams have properties that characterize the data, e.g., batch size or data type. 
For example, the capabilities, e.g., embedded CPU or GPU, of the edge devices that process the data have a strong impact on the performance. Similarly, the strategy for the service configuration ($c$) is relevant. Service configuration allows the node manager to set, or limit, the local computational resources, achieving elasticity in a constrained setting, but also impacting the SLOs' fulfillment; e.g., using a low-processing mode benefits energy efficiency, but limits the performance.
Therefore, we directly monitor the performance on the device, providing a continuous stream of metrics ($D$) for an immutable list of variables. 
%Finally, the output data is streamed to consumers, which is again characterized by a set of properties.
Figure~\ref{fig:stream-processing} summarizes this architecture, with its main components; in this work we focus on exploring the case of IoT sensors data stream.

% Given the properties and capabilities of different components in the stream processing architecture, the one that deserves more explanation is the service configuration:
% % , as changing the configurations will be our elasticity strategy for service adaptation.
% While streaming pipelines usually consist of a data provider, consumer, and the processing device in between, the edge device clearly presents the lowest latency for monitoring SLO fulfillment. At the same time, the processing device is the bottleneck for SLO fulfillment, because whether SLOs are fulfilled is primarily determined by the device capabilities and the service configuration.
% % The decision of where in the architecture to execute the AIF agent has substantial implications on its action scope, i.e., on the dynamic adaptations that takes to mitigate SLO violations.
% Hence, to minimize reaction time, the AIF agent is directly executed on the edge device. 
% % Given that we execute the agent on an edge device, the obvious choice is to adjust the local processing configuration.
% Whenever SLOs are violated, it is now the agent's responsibility to resolve this through elasticity strategies; however, while cloud computing can scale virtually unlimited amounts of processing resources, the edge device is restricted to its local scope. Hence, the preferred elasticity strategy of our AIF agents is to adjust service configurations. In the following, we introduce the model according to which the agent infers how to adjust the configuration.

\subsection{Generative Model Construction}

% \ins{misses what is a configuration}

Interpreting stream processing boils down to one central activity: reasoning about metrics. For this, we apply generative models, as introduced for AIF.
% thus, we facilitate the interpretation of metrics.
%In the following explanation of how to construct a generative model, we roughly adhere to 
Our implementation of the generative process follows the design by Parr et al.~\cite{parr_active_2022}, %This means, that we consider each of their design questions and explicitly answer how we solve them for a 
adapting it for the stream processing use case. Following~\cite{parr_active_2022}, we focus on the three main generative model design aspects:
\vspace{5pt}

\begin{enumerate}
    \item[\textbf{A}.] What are the generative model's \textit{components}, and what its \textit{interfaces}?
    \item[\textbf{B}.] What is the \textit{hierarchical} and \textit{temporal depth} of the generative model?
    \item[\textbf{C}.] What \textit{probabilities} are \textit{encoded} in the model, and how are they \textit{updated} over time?
\end{enumerate}
\vspace{4pt}

% \begin{table}[t!]
%   \centering
%   \caption{Model variables and their boundaries}
%   \label{tab:model-variables}
%   \begin{tabular}{lclc}
%     \toprule
%     Name   & Unit & Description &  Range  \\
%     \midrule
%     \textit{batch size}           & num        & number of machine parts per batch  & $[12, 30]$\\
%     \textit{utilization}          & \%        & utilization of the factory engine  & $[1, 100]$ \\
%     \textit{distance}             & cm        & space between two machine parts  & $[1, \infty[$\\
%     \textit{part delay}          & ms        & processing time per machine part  & $[1, \infty[$ \\
%     \textit{batch delay}          & ms        & total time for batch processing  & $[1, \infty[$ \\
%     \bottomrule
%   \end{tabular}
% \end{table}

\begin{figure}[t]
\centering
\subfloat[Variable relations]{\includegraphics[width=0.30\textwidth]{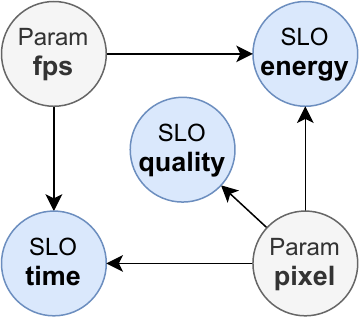}\label{subfig:dag}}
\hspace{0.03\textwidth}
\subfloat[Pragmatic value]{\includegraphics[width=0.33\textwidth]{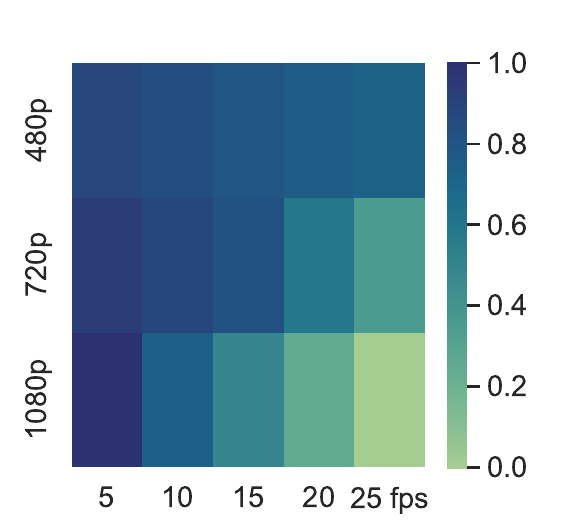}\label{subfig:gm-pv}}
\subfloat[Information gain]{\includegraphics[width=0.33\textwidth]{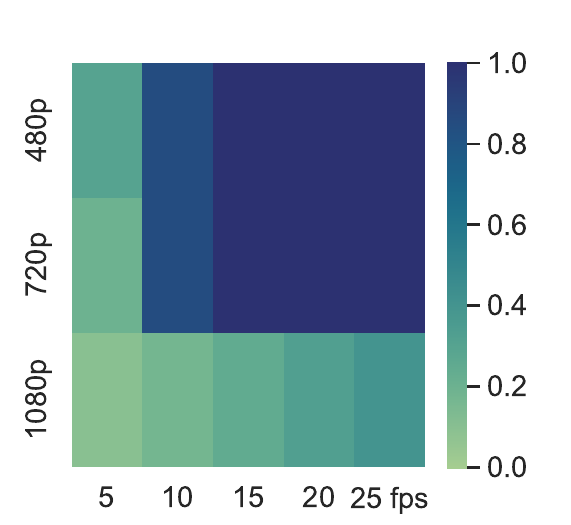}\label{subfig:gm-ig}}
\caption{Conditional (or causal) variable relations encoded in a Bayesian network; variable states (i.e., \textit{fps} and \textit{pixel}) form a 2D solution space where each parameter combination features a distinct pragmatic value (\textit{pv}) and information gain (\textit{ig})}
\label{fig:generative-model}
\end{figure}

%(A)
\textbf{A.} We envision a generative model consisting of two combined components: a directed graph that expresses variable relationships, and the precise conditional probabilities of which variable states are likely observed under certain environmental conditions. We represent the underlying graph as a Directed Acyclic Graph (DAG), where the edges indicate conditional dependency. For instance, consider Fig.~\ref{subfig:dag}, where the edge $\textit{fps} \rightarrow \textit{energy}$ shows how the number of frames per second (fps), i.e., a configuration parameter, has a decisive impact on the energy consumption of the processing device. If we specify an SLO for minimizing \textit{energy consumption}, such as $\textbf{energy} \leq 15 W$, it is possible to infer the probability of fulfilling it under different service configurations, i.e., according to the precise assignments of \textit{fps} and \textit{pixel}.

While applying the generative model appears straightforward, training it requires high-quality observations; one method to build it is through Bayesian Network Learning (BNL), e.g., as applied by~\cite{yazdi_resilience_2022,chen_causeinfer_2019,togacar_detecting_2022}. In the absence of training data, it is also possible to specify a BN through expert knowledge~\cite{kitson_survey_2023}. Given the resulting generative model, agents can reason how likely it is to observe SLO violations under a certain service configuration, e.g., exceeding $\textbf{energy}$ with $\textit{fps} = 15$. However, if the generative model does not reflect the generative process accurately, the agent can be surprised by the actual outcome. As defined by Dustdar et al.~\cite{dustdar_distributed_2023}, improving the model's accuracy requires a clear understanding of the \textit{interfaces} between the agent and the environment as these will map the observations from the generative process to internal elements of the generative model.

%(B)

% storyline: temporal depth one action into the future, evaluate every x ms

\vspace{2pt}
\textbf{B.} The temporal and hierarchical depth are two more sophisticated properties of the generative model. For the given use case, the temporal depth is bounded by the length of the policy $\pi$, i.e., how many steps in the future the behavior is predicted. Given that longer policies have a higher complexity for predicting the outcome, the policy length can be chosen in accordance with the evaluation frequency. This means, that for stream processing, it can be desirable to evaluate the SLO fulfillment with high frequency (e.g., every 500ms) and conversely keep the policy length low.

The hierarchical depth, on the other hand, is bounded by the number of variables and states in the generative model; for instance, Figures~\ref{subfig:gm-pv} and \ref{subfig:gm-ig} show a parameter space that consists of two variables, i.e., \textit{fps} and \textit{pixel}, that can take 5 and respectively 3 variable states. While this solution space and the DAG in Figure~\ref{subfig:dag} appears minimalistic, BNs can grow up to thousands of variables \cite{mengshoel_developing_2009}; depending on the granularity, it is possible to split up processes into increasingly smaller substructures, which can each be constrained and managed by a more specific set of SLOs~\cite{sedlak_diffusing_2024}. Hence, maintaining a desired level of abstraction~\cite{kirchhoff_markov_2018} ensures lower model complexity for inference and training. In that regard, extracting the MB around SLO variables~\cite{sedlak_equilibrium_2024} helps reduce the number of variables that must be considered.

% \ins{shrink and adapt:}
% For the given use case, we use an SLO-induced boundary as our natural limit on temporal depth: equal to the maximum \textit{batch delay (bd)}, each action-perception cycle lasts 500 ms. Within each cycle, the agent predicts the engine's behavior (i.e., reflected through the metrics) over the next 500 ms; afterward, the prediction is compared against the events observed during that time.
% While the cycle's length can be chosen freely, longer periods decrease the prediction accuracy or increase the computational complexity (i.e., to evaluate the SLO once, it must consider multiple cycles or fractions of them). The hierarchical depth, on the other hand, is determined by the number of variables and edges in the model. A deeper hierarchy would increase the complexity of model training and inference; however, the use case does not provide variables other than the ones already contained in the DAG.

%(C)
\vspace{2pt}
\textbf{C.} 
When an agent's observation presents a discrepancy with the expected outcome, e.g., an SLO is violated despite the previous agent intervention to have a proper configuration, it generate a ``surprise,'' calling the agent to adjust its generative model. Here, the priors are both the variable relations and the conditional probabilities encoded in the generative model. This means, that upon observing something that goes contrary to its expectation, the agent updates its model to form its posterior beliefs. While variable relations and conditional probabilities are subject to retraining, we assume the list of monitored variables to be immutable. Nevertheless, given that feature-evolving streams are gaining popularity~\cite{chen_novel_2024}, extending this to mutable lists of variables promises a valuable asset. 

For the generative model presented in this paper, we distinguish two types of variables: 
% Variables have substantial differences in the way that they determine stream processing: 
(1) the configuration parameters (i.e., parents in Figure~\ref{subfig:dag}), which decide \textit{how} streaming data is processed, and (2) the SLOs (i.e., child nodes), which constrain \textit{whether} the stream processing behavior fulfills the agent's preferences. 
The classification into parameters stems from the system definition, hence, it is immutable and determined depending on which variables the agent can adjust locally. However, for remaining variables, the decision to turn them into SLOs, can also be taken at a later stage. Thus, it is possible to adjust the list of SLOs and, more importantly, also their desired thresholds. For instance, depending on the type of processing device, we can decide to cap \textbf{energy} either by 10W or 15W. Thereby, we described the characteristics and boundaries of the generative model as required to interpret continuous stream processing.

\subsection{Active Inference Cycle}

To continuously ensure model accuracy during stream processing, and consequently high SLO fulfillment, we embed the generative model into the action-perception loop executed by an AIF agent. 
\begin{figure}[t]
\centerline{\includegraphics[width=1.0\textwidth]{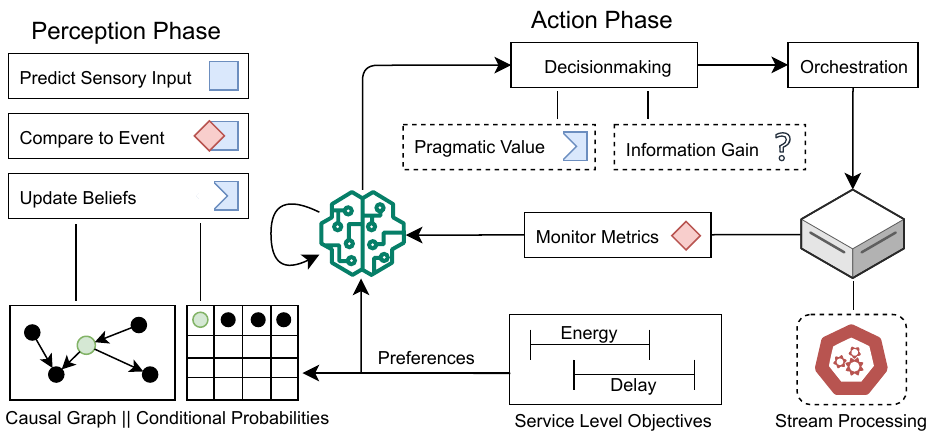}}
\vspace{5pt}
\caption{High-level action-perception cycle in AIF; stream processing metrics are interpreted using a generative model; the agent updates the model according to prediction errors and makes adjustments by balancing pragmatic value with information gain}
\label{fig:action-perception}
\end{figure}
Figure~\ref{fig:action-perception} gives a high-level overview of this action-perception loop: a stream processing service (red) is executed on an edge device; during this execution, the monitored metrics are provided to the AIF agent. Application stakeholders (e.g., DevOps engineers) can specify SLOs that should be ensured during processing; these SLOs constitute the agent's preferences of what it would like to observe. As the AIF agent receives a batch of metrics (i.e., observations), it evaluates the degree to which SLOs were fulfilled according to Eq.~\eqref{eq:slo-f}:
\begin{subequations}
\label{eq:slo-f}
    \begin{equation}
        \label{eq:Q-phi}
        \phi(Q) = \frac{\sum_{i = 1}^{|Q|} \phi(q_{i})}{|Q|}
    \end{equation}    
    \vspace{-4pt}
    \begin{equation} 
        \label{eq:q-phi}
        \phi(q_{i}) = \phi(q_{i}, d \mid \forall d \in {D_{q_{i}}}) =
            \sum_{j=1}^{|D_{q_{i}}|}{
                \frac{\phi(d_{j}, q_{i})}{|D|}
            }
    \end{equation}
    \vspace{4pt}
    \begin{equation}
        \label{eq:q-phi-m}
        \text{where }\phi(q_{i}, d_{j}) = 
        \begin{cases}
        1,& \text{if}\ d^{q_{i}}_{j_{\text{min}}} \leq d_{j} \leq d^{q_{i}}_{j_{\text{max}}} \\
        0,& \text{otherwise}
        \end{cases}
        \vspace{8pt}
    \end{equation}
\end{subequations}

\noindent where $Q$ represents a set of SLOs, $D$ the list metrics, and $\phi$ the resulting SLO fulfillment. As shown in Eq.~\eqref{eq:q-phi-m}, the overall SLO fulfillment is determined by the ratio of samples that are found in the desired range, delimited by the thresholds. The actual SLO fulfillment is then compared to the agent's prediction according to its generative model; when the two diverge, the agent adjusts the generative model, which means retraining the causal graph and the variables' conditional probabilities. 

To minimize EFE through optimal action, agents consider the known concepts of either (1) changing the environment toward its preferences to maximize pragmatic value (\textit{pv}) or (2) improving decision-making by resolving contextual information to maximize information gain (\textit{ig}). In the following, we introduce how we calculate a simplified version of the information gain as shown below in Eq.~\eqref{eq:ig-simplified}:
\begin{subequations}
\begin{equation}
    \Im (D \mid m) = \sum_{i=1}^{|D|} - \log P(d_i \mid m)
    \label{eq:surprise-batch}
\end{equation}
\begin{equation}
    ig (c) = \left(\frac{\tilde{\Im}_{c}}{\tilde{\Im_\omega}}\right) \times 100
    \label{eq:ig}
    \vspace{8pt}
\end{equation}
\label{eq:ig-simplified}
\end{subequations}
where the \textit{ig} of a configuration \textit{c} is calculated based on the surprise that this configuration caused in the past. 
If a configuration repeatedly shows surprising results, this indicates that the agent cannot yet predict its \textit{pv} accurately; consequently, this assigns a high \textit{ig} to it.
While Eq.~\eqref{eq:surprise} provided the idea of how to calculate the surprise ($\Im$) for a single observation, Eq.~\eqref{eq:surprise-batch} shows how the surprise for a batch of metrics equals the sum of the individual metrics. Hence, when calculating the \textit{ig} for a configuration, e.g., $\textit{fps} = 15$ and $\textit{pixel} = 480$ in Figure~\ref{subfig:gm-ig}, the agent considers how the median surprise ($\tilde{\Im}_{c}$) for the configuration relates to the global median surprise ($\tilde{\Im_\omega}$). The weights that the agent assigns to \textit{pv} and \textit{ig} are subject to hyperparameter optimization; to speed up convergence, we assign \textit{pv} double the weight of \textit{ig}, i.e., $w_{pv} = 2 \times w_{ig}$.

Up to this point, explanations of \textit{pv} and \textit{ig} were built on the assumption that the generative model can always provide an approximate probability for any possible state. However, the generative model might not be able to provide this approximation without having observations for a certain parameter configuration. Further, recall that we are dealing with discrete states, thus, even though an agent infers based on observations how likely it is to fulfill SLOs with $\textit{fps} = \{10,20\}$, this lacks understanding of what will happen if $\textit{fps} = 15$. Hence, we interpolate between the \textit{pv} and \textit{ig} of the closest neighbors, or for the first initial values, choose the closest neighboring value. Notice, how the interpolation took place in Figures~\ref{subfig:gm-pv} and \ref{subfig:gm-ig}; this becomes particularly important because the agent in its simplest form minimizes EFE (i.e., \textit{pv} and \textit{ig}) by comparing a potentially exponential number of parameter combinations. While the combinatorial complexity requires dedicated optimization, the interpolation between configurations provides a workaround in cases when no observations are available.

As soon as the agent has come to a decision, it has to orchestrate the action, i.e., by adjusting the local processing configuration. Thus, the action-perception cycle is closed; whenever the agent receives a new batch of observations, it repeats this cycle by evaluating the actual SLO fulfillment, adjusting its generative model according to the new observations, and using its generative model to infer a service adaptation.

\section{Evaluation}
\label{sec:analysis}

% \ins{introduction to the evaluation, also explain on a high-level what we want to evaluate}

In the following section, we present the implementation and evaluation of our methodology over a heterogenous set of actual services and real edge devices. To underline the generality of our approach, we showcase how our stream processing framework fits multiple use cases; the overarching goal is to evaluate whether AIF agents can dynamically find a satisfying stream processing configuration. For each use case, we document the experimental setup, including the service implementations and applied processing hardware. Finally, we present the experimental results and conclude with a critical discussion.

\subsection{Implementation}
\label{subsec:implementation}

% \ins{description of implementation: prototype, service execution environment, training data \& pgmpy, hyperparameters}

To embed the designed AIF agent into actual processing devices, we provide a Python-based prototype that comprises both the generative model construction as well as the continuous action-perception cycle. The prototype of the AIF agent, the implementation of the processing services, and the experimental results are all shared in one publicly accessible repository\footnotemark. Thus, we aim to make our results reproducible and allow others to interact with our methodology.
\footnotetext{The prototype of the stream processing framework is available at \href{https://github.com/borissedlak/intelligentVehicle/tree/master/ES_EXT}{GitHub}, accessed on July 31st 2024}

To isolate resource consumption, we execute each processing service in a Python thread. During that time, each service observes its own SLO fulfillment as part of its action-perception cycle; in the present state, this is done every 2000ms, though it can be customized for different service types.
To avoid interfering with the stream processing task, the training of the generative model is detached from the processing thread.
To train and update its generative model, AIF agents use pgmpy~\cite{ankan_pgmpy_2023}, a Python library for Bayesian Network Learning (BNL). As shown in previous work~\cite{sedlak_designing_2023,sedlak_equilibrium_2024}, pgmpy is very effective for providing both the structure and the conditional dependencies of BNs, making it apt for the continuous action-perception cycle.
% In pgmpy, BNs can be encoded in XML, which each had a size of roughly 10kB in our evaluation; hence, a feasible size to be transmitted and shared within the platoon.

\subsection{Experimental Setup}
\label{subsec:experimental-setup}

% \ins{explain the services and the devices, the power modes and the impact on the device}

To evaluate our prototype we provide three use cases for stream processing, showing for each of them how our framework can ensure its processing requirements. Table~\ref{tab:service-list} provides essential information on the three processing services, such as the number of configuration parameters and SLOs. In the following, we describe each of them in more detail, including their practical relevance and application:

\begin{itemize}
    \item[\textbf{CV.}] Videos rank among the most commonly streamed data types; hence, Computer Vision (CV) gained increasing popularity for detecting objects in videos~\cite{yi_lavea_2017} or ensuring privacy~\cite{sedlak_privacy_2023} by transforming the video content. Hence, the first processing service is a video inference task based on Yolov8~\cite{varghese_yolov8_2024} that detects vehicles, traffic lights, and pedestrians in a traffic junction. 
    % REVISION: Include other privacy paper as well
    
    \item[\textbf{LI.}] Sticking to road traffic and transportation, autonomous driving is empowered by multitudes of sensors embedded in Autonomous Vehicles (AVs). As such, Lidar is commonly used to scan AVs' surroundings through point cloud processing~\cite{liu_point_2021}; AVs use this to react in real-time to dynamic traffic conditions. Hence, the second service processes point clouds through Lidar (LI) using the SFA3D~\cite{dzung_maudzungsfa3d_2020} library.
    
    \item[\textbf{QR.}] Within recent years, QR codes have quickly found their way into our everyday lives, e.g., to share contact information or hyperlinks. However, QR codes are also applied to track objects~\cite{ahmadinia_tracking_2023}, i.e., by attaching the code to an object and continuously inferring its position. Hence, the third service uses OpenCV~\cite{opencv_opencv_2024} to track the position of a vehicle according to a QR code attached to its rear.
\end{itemize}

\begin{figure}[t]
\centering
\subfloat[CV (Yolov8)]{\label{subfig:demo-cv}\centering\includegraphics[width=.325\columnwidth]{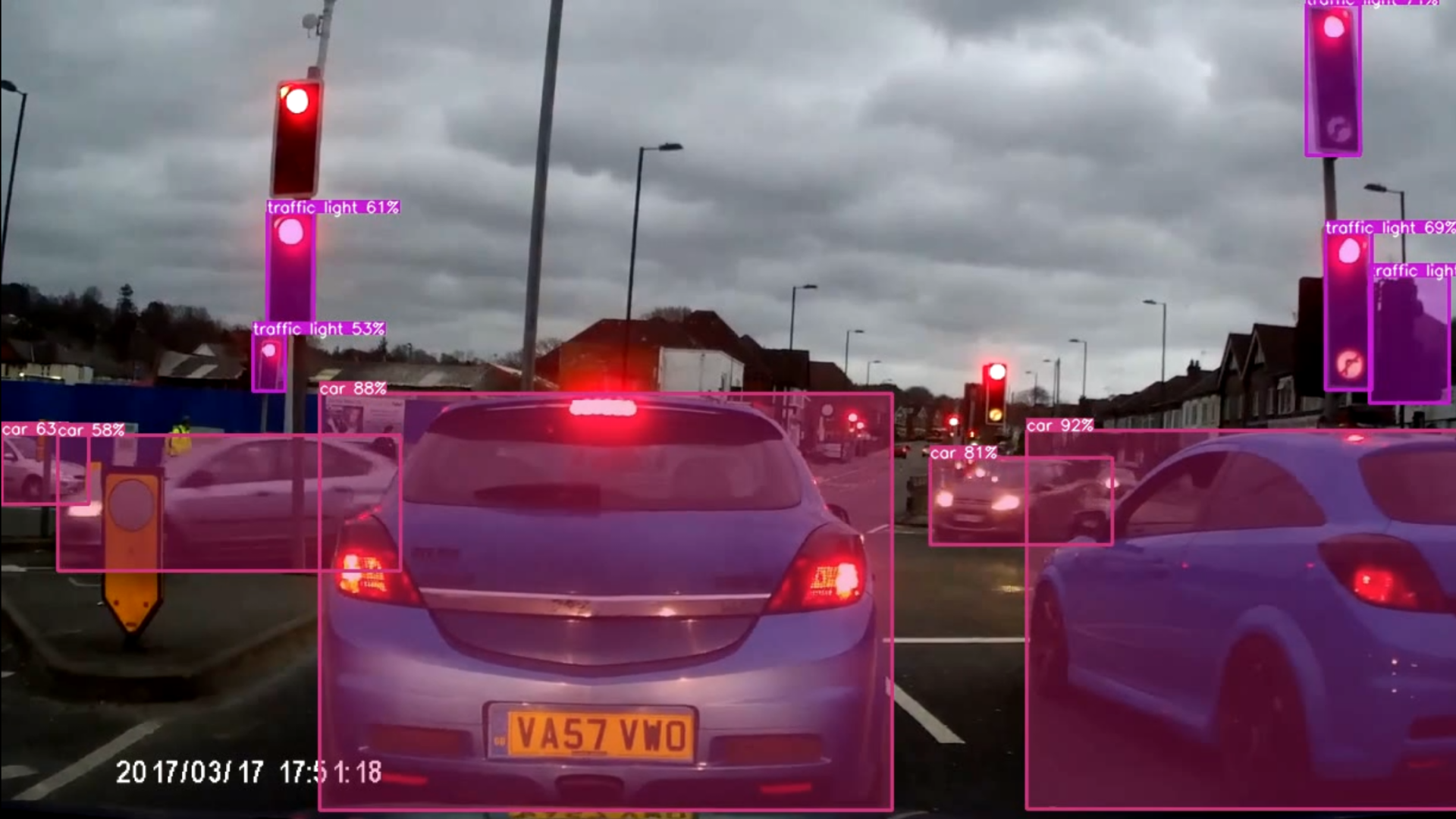}}
\hspace{0.005\columnwidth}
\subfloat[QR (OpenCV)]{\label{subfig:demo-li}\centering\includegraphics[width=.314\columnwidth]{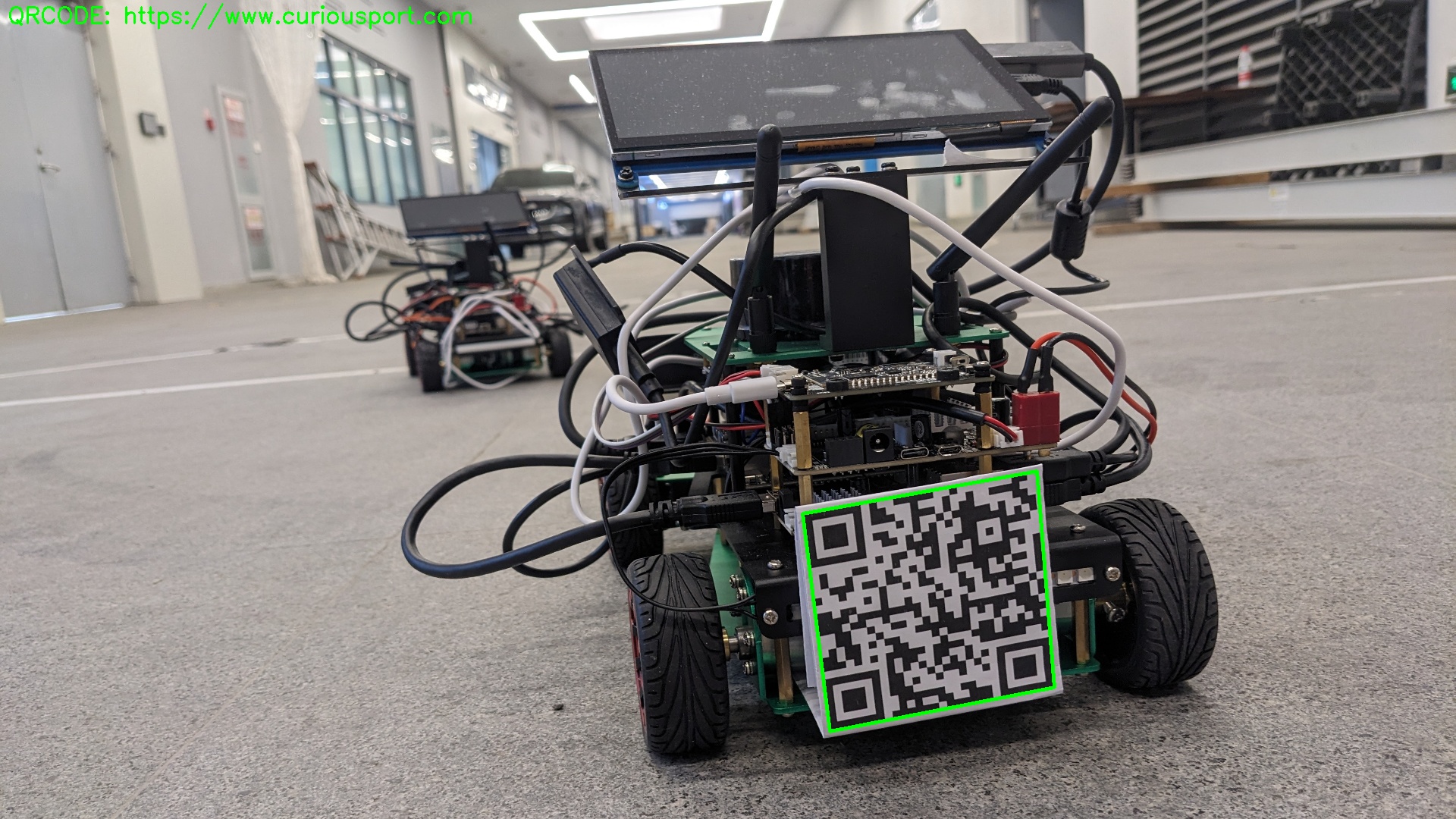}}
\hspace{0.005\columnwidth}
\subfloat[LI (SFA3D)]{\label{subfig:demo-li}\centering\includegraphics[width=.324\columnwidth]{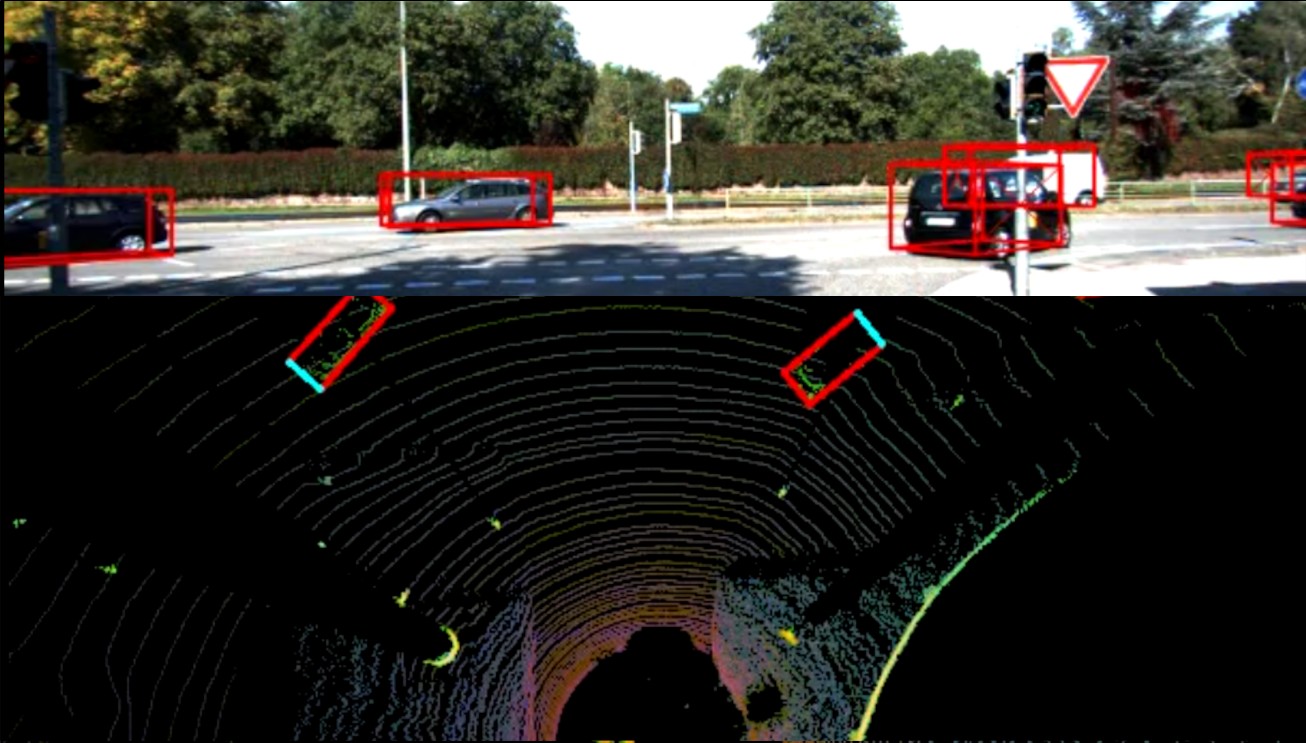}}
\caption{Demo output for each service according to prerecorded input data; all three services are iteratively processing video streams (CV and QR) or binary input (LI)}
\label{fig:services-demo}
% \vspace{-15pt}
\end{figure}

Figure~\ref{fig:services-demo} shows a demo output for each service to make more tangible what contents are processed and produced by the three services.
Notice, that to ensure a stable evaluation environment, the service processed either prerecorded videos (\textit{CV} \& \textit{QR}) or binary-encoded point clouds ($LI$).
For adjusting the outcome, each service has specific configuration parameters available, such as the resolution (\textit{pixel}) and \textit{fps} for \textit{CV} and \textit{QR}; \textit{LI} accepts an additional parameter \textit{mode} to define the point cloud radius. While \textit{pixel} and \textit{fps} are common properties of video streams, we assume that the AIF agent can still change these during processing, i.e., by shrinking the maximum video resolution or sampling the video to a lower frame rate.  

We empirically map each service's expected QoS level to a list of SLOs, based on our expert knowledge. After several experiments, which are out of the scope of this work, the following values proved especially useful: we constrain the processing \textbf{time} to $\leq 1000 / \textit{fps}$ (i.e., frames must be processed faster than they come in). To ensure efficient stream processing, the maximum \textbf{energy} consumption is capped at $\leq 15 W$. 
For what concerns the video resolution (\textit{pixel}) provided to \textit{CV}, the service uses the respective Yolov8 model size (i.e., v8n, v8s, v8m). However, this affects the number of objects that are detected, which is ensured through the \textbf{rate} SLO. The presented configuration parameters and SLOs describe the list of variables that will be included in the generative model; nevertheless, only through BNL it is possible to extract their relations and dependencies.

\setlength{\tabcolsep}{5pt}
\begin{table}[t]
    \small
  \centering
  \caption{List of all stream processing services covered by the framework}
  \label{tab:service-list}
  \begin{tabular}{clccc}
    \toprule
    ID & Service Description & CUDA & Parameters & SLOs\\
    \midrule

    \textit{CV} & Object Detection with Yolov8~\cite{varghese_yolov8_2024} & Yes & \textit{pixel}, \textit{fps} &  \textbf{time}, \textbf{energy}, \textbf{rate}\\
    \textit{LI} & LiDAR Point Cloud Processing~\cite{dzung_maudzungsfa3d_2020} & Yes & \textit{mode}, \textit{fps} & \textbf{time}, \textbf{energy} \\
    \textit{QR} & Detect QR Code w/ OpenCV~\cite{opencv_opencv_2024} & No & \textit{pixel}, \textit{fps} & \textbf{time}, \textbf{energy} \\
    \bottomrule
  \end{tabular}
  % \vspace{-10pt}
\end{table}

We evaluate on two different instances of Nvidia Jetson boards, namely Jetson Orin \textit{NX} and Orin \textit{AGX}, which are described in more detail in Tab.~\ref{tab:device-list}. Since these devices have different processing capabilities, local AIF agents cannot infer parameter configurations using one global model because the mismatch between generative model and device will likely cause the agent to be surprised. Hence, this device heterogeneity requires AIF agents to train their generative models separately.
Further, to decrease energy consumption, Jetson boards offer the feature of operating in different device modes, which limits the available device resources. However, this introduces further heterogeneity, even between identical devices operating in different modes; hence, the list of devices contains for both devices two entries: $AGX_+$ indicates that the device operated without constraints, and $AGX_-$ indicates that the local resources were limited. 
Notice, how switching between device modes reduces the number of CPU and GPU cores for both device types.

While the Jetsons' specific version of Nvidia CUDA has only minor importance, CUDA itself is crucial to speed up video or point cloud processing through GPU acceleration. However, to the best of our knowledge, there are no off-the-shelf solutions to use CUDA for scanning QR codes with OpenCV; hence, CUDA was only used to accelerate \textit{CV} and \textit{LI}, while \textit{QR} was processed entirely on the CPU.
% Each Jetson \textit{NX} is embedded in a Rosmaster R2\footnotemark car -- a battery-powered multi-sensory vehicle used for development. 

\setlength{\tabcolsep}{4.5pt}
\begin{table}[t]
    \small
  \centering
  \caption{List of edge devices involved in the evaluation; depending on the Jetson power mode, devices have different numbers of CPU and GPU cores available}
  \label{tab:device-list}
  \begin{tabular}{lccccrcr}
    \toprule
    ID & Full Device Name & Mode & Price\footnotemark & CPU & RAM &  GPU & CUDA \\
    \midrule

    $AGX_+$ & Jetson Orin AGX & MAX & 800 €  & ARM 12C  & 64 GB  & Volta 8k & 12.2\\
    $AGX_-$ & Jetson Orin AGX & LIM & 800 €  & ARM 8C  & 64 GB  & Volta 4k & 12.2\\
    $NX_+$ & Jetson Orin NX  & MAX & 450 €  & ARM 8C  & 8 GB  & Volta 4k & 11.4 \\
    $NX_-$ & Jetson Orin NX  & LIM & 450 €  & ARM 4C  & 8 GB  & Volta 2k & 11.4\\
    % ThinkPad X1 G10  & \textit{Lapt.} & 950 € & Intel i7-1260P & 32 GB & ------ & ------\\
    \bottomrule
  \end{tabular}
  % \vspace{-10pt}
\end{table}
\footnotetext{Prices adopted from \href{https://sparkfun.com/}{sparkfun}, accessed Jul 31st 2024}

\subsection{Results \& Discussion}
\label{subsec:experimental-setup}

Given the experimental setup, we evaluate the prototype based on: 
(1) how long does the AIF agent to find a satisfying parameter configuration;
(2) is the behavior of the AIF agent explainable during its action-perception cycle;
(3) how does the service and device heterogeneity impact agents' decisions.
For this, we conduct a total number of 12 experiments, which stems from the fact that we evaluate each services on each device type. During the experiments, we capture the current SLO fulfillment, the preferred configuration, and the \textit{pv} and \textit{ig} values that the AIF agent assigns to each parameter combination. Additionally, we collect the structure of the generative model.

\begin{figure}[t]
\centering
\subfloat[CV (Yolov8)]{\includegraphics[width=0.33\textwidth]{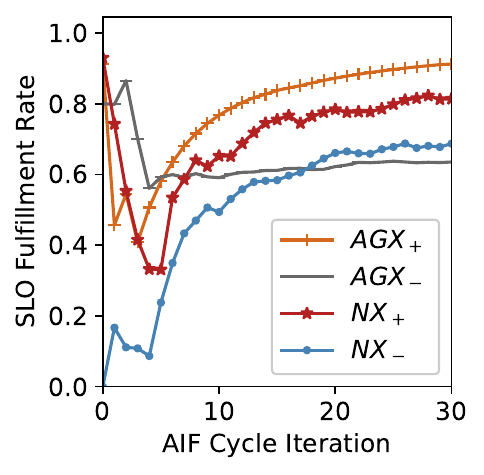}\label{subfig:slo-f-CV}}
\subfloat[QR (OpenCV)]{\includegraphics[width=0.33\textwidth]{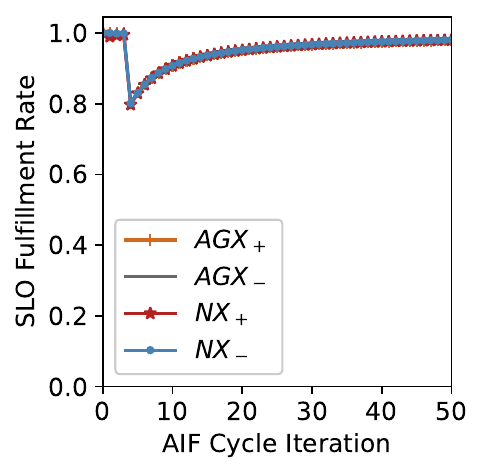}\label{subfig:slo-f-QR}}
\subfloat[LI (SFA3D)]{\includegraphics[width=0.33\textwidth]{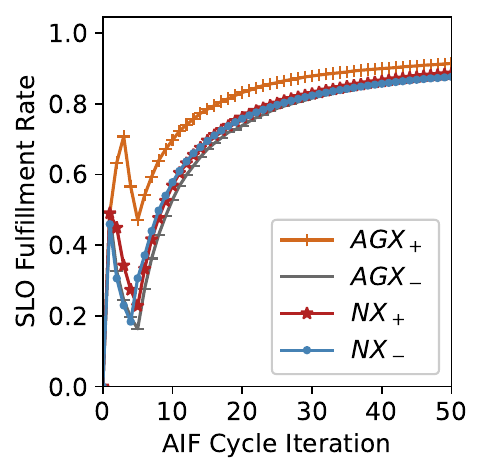}\label{subfig:slo-f-LI}}
\caption{Empirical SLO fulfillment measured during service execution; initial training rounds show unstable behavior, while later rounds converge to a clear preference }
\label{fig:slo-f}
\end{figure}

\begin{table}[t!]
  \centering
  \caption{Preferred configuration per service x device combination\vspace{5pt}}
  \label{tab:preferred-configs}
  \begin{tabular}{r|ccc}
    % \toprule
     & \textit{CV} & \textit{QR} &  \textit{LI}  \\
    \midrule
    $AGX_+$           &  $\langle 1080\text{p}, 5\text{ fps} \rangle = 0.94$ & $\langle 720\text{p}, 15\text{ fps} \rangle = 1.0$ & $\langle \text{single}, 5\text{ fps} \rangle = 0.98$ \\
    $AGX_-$          & \cellcolor{orange!15} $\text{}\langle 720\text{p}, 15\text{ fps} \rangle = 0.62 $ & $\text{   }\langle 720\text{p}, 5\text{ fps} \rangle = 1.0$  & $\langle \text{single}, 5\text{ fps} \rangle = 0.93$\\
    $NX_+$             & $\langle 720\text{p}, 10\text{ fps} \rangle = 0.83$  & $\text{   }\langle 720\text{p}, 5\text{ fps} \rangle = 1.0$ & $\langle \text{single}, 5\text{ fps} \rangle = 0.92$\\
    $NX_-$          & $\text{  }\langle 480\text{p}, 5\text{ fps} \rangle = 0.73$ & $\langle 480\text{p}, 10\text{ fps} \rangle = 1.0$ & $\langle \text{single}, 5\text{ fps} \rangle = 0.90$  \\
    \bottomrule
  \end{tabular}
\end{table}

Figure~\ref{fig:slo-f} shows the SLO fulfillment that the AIF agents reported during each of its iterations. For each of the services, the tendency is that the agent initially reports low SLO fulfillment while it is still exploring the solution space; however, after 20-30 iterations the agent converges to a clear preference. Table~\ref{tab:preferred-configs} shows the respective parameter configurations to which the agents converged. For 11 out of 12 cases, we could verify through exhaustive comparison that the chosen parameter configuration was optimal, or showed only minor divergence. For the \textit{CV} service on $AGX_-$ (grey line in Figure~\ref{subfig:slo-f-CV}), however, the exploration did not provide the desired results, and so it converged to a sub-optimal configuration.

Given this, we conclude that the AIF agent is not guaranteed to find the optimal solution; however, the results showed that in most of the cases it was still possible to find the optimal solution. Hence, using the AIF agent to supervise the processing service showed promising results. Combined with its low processing overhead~\cite{sedlak_equilibrium_2024}, this makes it a great fit for for resource-constrained edge devices.

\begin{figure}[t]
\centering
\subfloat[CV (Yolov8)]{\includegraphics[width=0.33\textwidth]{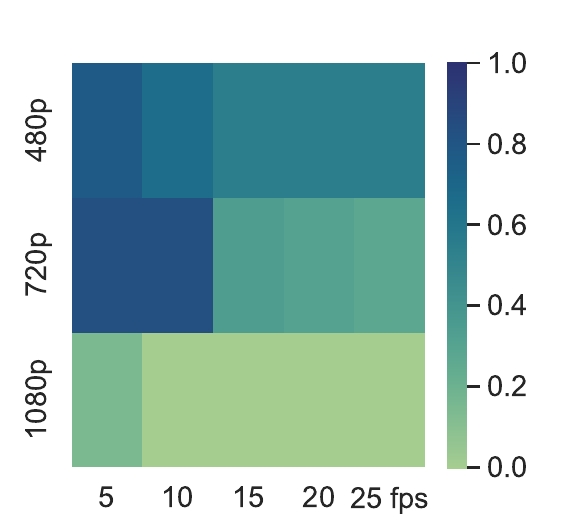}\label{subfig:pv-matrix-CV}}
\subfloat[QR (OpenCV)]{\includegraphics[width=0.33\textwidth]{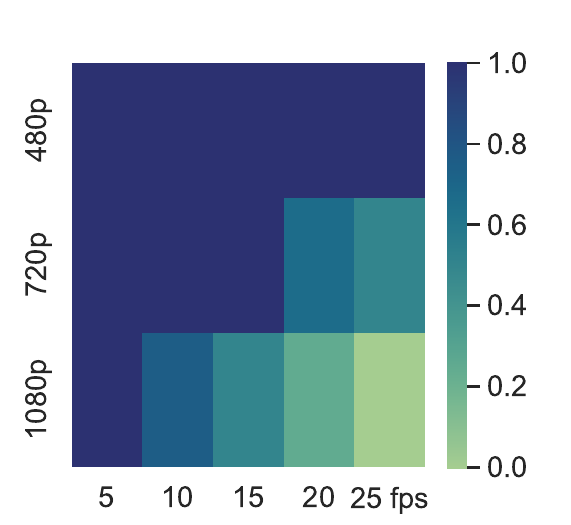}\label{subfig:pv-matrix-QR}}
\subfloat[LI (SFA3D)]{\includegraphics[width=0.33\textwidth]{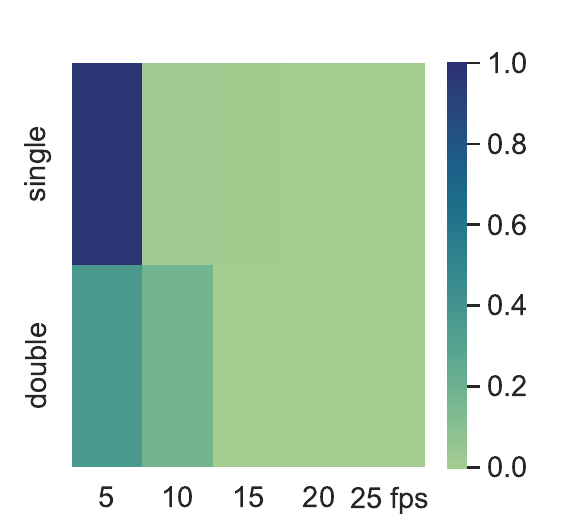}\label{subfig:pv-matrix-LI}}
\caption{PV Matrices for the three services executed at $NX_{+}$; individual cells in the heatmap show the expected SLO fulfillment for each combination of config parameters}
\label{fig:pv-matrices}
\end{figure}

\begin{figure}[t]
\centering
\subfloat[CV (Yolov8)]{\includegraphics[width=0.33\textwidth]{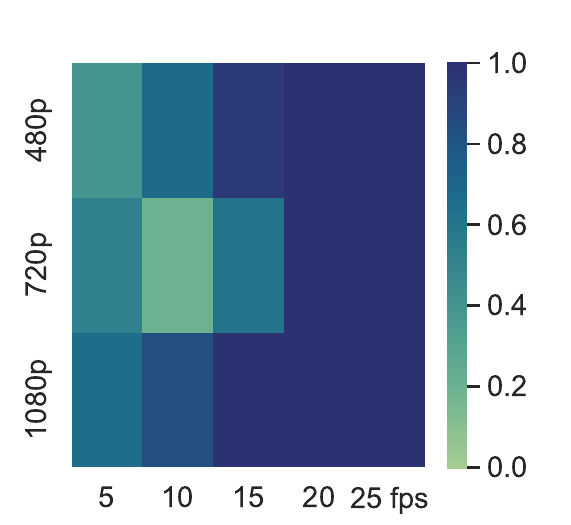}\label{subfig:ig-matrix-CV}}
\subfloat[QR (OpenCV)]{\includegraphics[width=0.33\textwidth]{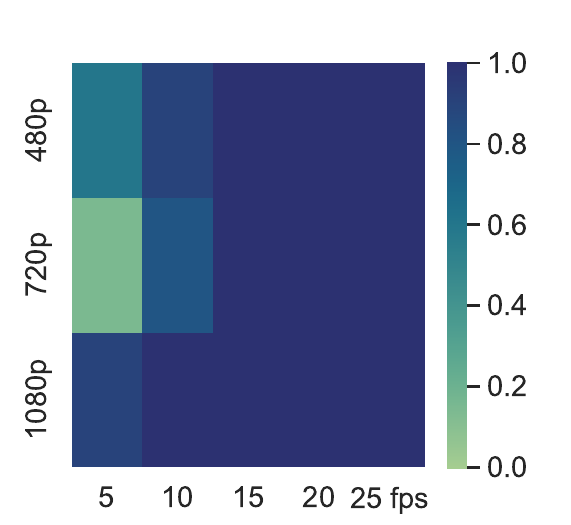}\label{subfig:ig-matrix-QR}}
\subfloat[LI (SFA3D)]{\includegraphics[width=0.33\textwidth]{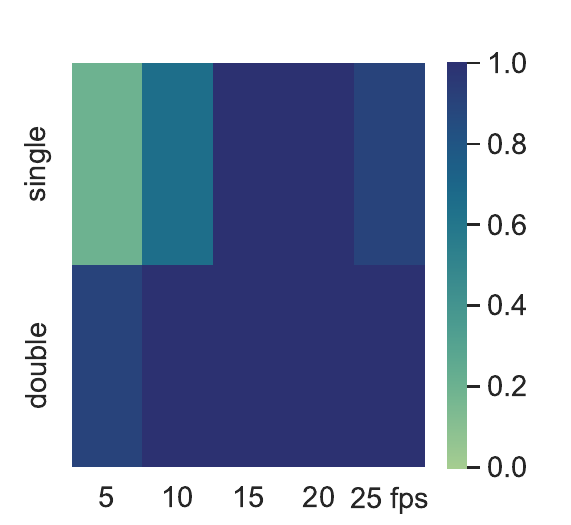}\label{subfig:ig-matrix-LI}}
\caption{IG Matrices for the three services executed at $NX_{+}$; individual cells in the heatmap indicate expected model improvement when using the specific configuration}
\label{fig:ig-matrices}
\end{figure}

\begin{figure}[t]
\centering
\subfloat[CV (Yolov8)]{\includegraphics[width=0.28\textwidth]{figure2a.pdf}\label{subfig:dag-CV}}
\hspace{0.04\textwidth}
\subfloat[QR (OpenCV)]{\includegraphics[width=0.28\textwidth]{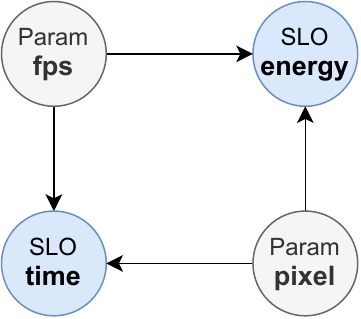}\label{subfig:dag-QR}}
\hspace{0.04\textwidth}
\subfloat[LI (SFA3D)]{\includegraphics[width=0.28\textwidth]{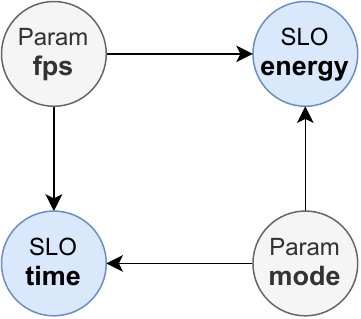}\label{subfig:dag-LI}}
\caption{Directed Acyclic Graphs (DAGs) reflecting the variable relations in the generative models; relations were extracted through BNL and are empirically verifiable}
\label{fig:result-dags}
\end{figure}

To analyze whether the behavior of the AIF agent is explainable, we retrace its steps and verify how it selects its preferred parameter combinations. In Figures~\ref{fig:pv-matrices} and \ref{fig:ig-matrices}, we provide the final matrices for the \textit{pv} and \textit{ig} values that the agent assigned to each parameter configuration at the end of the experiment. Recall, that in case a configuration was not empirically evaluated yet, the agent interpolates its \textit{pv} and \textit{ig} according to neighboring configurations. 
Given the figures, we notice that the matrices are coherent with the preferred configurations in Table~\ref{tab:preferred-configs}: the chosen configuration showed both a high \textit{pv} due to the expected SLO fulfillment, but also a low \textit{ig} due to the increasing exploitation, i.e., the configuration promise little model improvement.
Additionally, Figure~\ref{fig:result-dags} shows the structure of the generative models for each of the processing services, that is, how the AIF agent linked the configuration parameters and SLO variables during BNL. It is according to this variable relations, that the agent reasons how it must adjust \textit{pixel} and \textit{fps} to fulfill the SLOs.

Given that, we conclude that the agents' behavior is rationally explainable because the edges in the DAG are logical and match our expert knowledge. Using this graph, it is possible to explain why the agent converged to a certain configuration, e.g., \textit{CV} at $NX_+$ to $\textit{pixel} = 5$ and $\textit{fps} = 5$, because this configuration showed clearly the highest \textit{pv} value in Figure~\ref{subfig:pv-matrix-CV}. Still, although its \textit{ig} value in Figure~\ref{subfig:ig-matrix-CV} was lower, the agent decided that no further exploration was needed. Notice how in both Figures~\ref{subfig:pv-matrix-CV} and \ref{subfig:pv-matrix-QR}, there are multiple parameter configurations that promise equal \textit{pv} values; however, the agent randomly selects one of them because they all fulfill its preferences equally. Given that we would specify an SLO that minimizes \textit{energy}, not just cap it, this would lead to a more specific preference here. 
Together, these rules form coherent patterns that can be empirically verified, thus increasing the trust in agents' results.

To answer the third research question, we use the existing results. Given those, we conclude that the presented framework for adaptive stream processing showed promising results for all combinations of heterogeneous devices and services. In particular, Table~\ref{tab:preferred-configs} underlines how the agents adjusted their local generative model to find configurations that match their device capabilities, e.g., $AGX_+$ chose $\textit{pixel} = 1080p$ for \textit{CV} and hence optimized the \textbf{rate} SLO, whereas $NX_-$'s capabilities only sufficed to choose $\textit{pixel} = 480p$. Given all that, we see strong potential for using the agent to optimize the QoS of other stream processing use cases; the requirements for this are known from Section~\ref{sec:agent-design}: observable stream processing and clear variable specifications, i.e., parameters and SLOs. However, to make this framework apt for more complex scenarios, further evaluations are needed that operate with larger solution spaces.

\section{Conclusion}
\label{sec:conclusion}

This paper presented a novel framework for adaptive stream processing that continuously ensures QoS during processing on resource-constrained edge devices. Processing services were supervised through Active Inference, a behavioral framework from neuroscience, which allows agents to develop and maintain a logical model of how to ensure processing requirements, also called SLOs. Thus, edge devices can ensure that stream processing complies with requirements. 
% This approach does not require an initial data set to train ML models -- rather it trains the model incrementally according to new observations, thus canceling our data drifts. 
By executing AIF agents directly on edge devices, it is possible to observe with low latency how different service configurations impact SLO fulfillment. Depending on the outcome, AIF agents continuously adjust their generative model and use it to infer service configurations that promise high SLO fulfillment or further model improvement. To that extent, the AIF agents make use of causal variable relations that determine how their actions affect SLO fulfillment; this increases the trustworthiness of inferred configurations.

To evaluate the framework, we implemented a Python-based AIF agent for monitoring and guaranteeing SLO fulfillment, by adapting local processing configurations. In this way, we implemented an elasticity strategy on constrained edge devices, that are otherwise unable to scale resources as in cloud computing scenarios. We evaluated our framework on three different scenarios for stream processing, two for video processing and one for Lidar sensors, which were executed on heterogeneous edge devices. Not only did the devices, i.e., two instances of Nvidia Jetson, have different processing capabilities, but they could also operate in different hardware modes. Given this setup, the problem was to find service configurations that fulfill SLOs on processing time, energy consumption, and detection rate. Our results showed that the AIF agents required roughly 20 to 30 iterations in the action-perception cycle to converge to the optimal solution; further, the inferred results were empirically verifiable and followed logical patterns.
%
% we conducted a design study that optimizes the throughput for a smart manufacturing use case. Which action the agent takes and how it adapts its beliefs was determined by three main factors: pragmatic value, assigned risk (of violating SLOs), and information gain. We implemented the AIF agent in Python and tracked each cycle's preferred action -- including the factors that led to it -- and the agent's causal understanding between two variables.
% After 5 cycles, the agent converged to a solution that presented an optimal tradeoff between high pragmatic value and negligible SLO violations. Further, the agent needed only 30 observations (i.e., 2 cycles) to estimate a previously unknown variable relation. Exploring causalities between variables and constructing the agent's behavior from empirical factors makes the produced solutions traceable. 
Based on that, we see strong potential for AIF agents to support elastic stream processing in further, real time scenarios.
We plan to extend the evaluation of our contribution to more complex scenarios, where the solution space is larger. Furthermore, we plan to have an in depth comparison of how AIF performs when compared to other ML methods.

\section*{Funding}
This work has been supported by the European Union's Horizon Europe research and innovation program under grant agreements No. 101135576 (INTEND) and No. 101070186 (TEADAL).

\section*{Data availability}
The datasets used in this article are cited appropriately.

\section*{Conflict of interest}
Authors declare that there are no conflicts of interest.

\bibliography{Boris, Andrea, Victor}

\end{document}